\def\year{2019}\relax
\documentclass[letterpaper]{article} 
\usepackage{aaai19}  
\usepackage{times}  
\usepackage{helvet}  
\usepackage{courier}  
\usepackage{url}  
\usepackage{graphicx}  
\usepackage{amsmath}
\usepackage{amssymb}
\usepackage{amsthm}
\usepackage{enumerate}
\usepackage{multirow}
\usepackage{caption}
\usepackage{subfigure}
\usepackage{balance}
\begin{document}
%
\title{Agile Domain Adaptation}
\author{Jingjing Li, Mengmeng Jing, Yue Xie, Ke Lu and Zi Huang}

\maketitle
\begin{abstract}
Domain adaptation investigates the problem of leveraging knowledge from a well-labeled source domain to an unlabeled target domain, where the two domains are drawn from different data distributions. Because of the distribution shifts, different target samples have distinct degrees of difficulty in adaptation. However, existing domain adaptation approaches overwhelmingly neglect the degrees of difficulty and deploy exactly the same framework for all of the target samples. Generally, a simple or shadow framework is fast but rough. A sophisticated or deep framework, on the contrary, is accurate but slow. In this paper, we aim to challenge the fundamental contradiction between the accuracy and speed in domain adaptation tasks. We propose a novel approach, named {\it agile domain adaptation}, which agilely applies optimal frameworks to different target samples and classifies the target samples according to their adaptation difficulties. Specifically, we propose a paradigm which performs several early detections before the final classification. If a sample can be classified at one of the early stage with enough confidence, the sample would exit without the subsequent processes. Notably, the proposed method can significantly reduce the running cost of domain adaptation approaches, which can extend the application scenarios of domain adaptation to even mobile devices and real-time systems. Extensive experiments on two open benchmarks verify the effectiveness and efficiency of the proposed method.
\end{abstract}

\section{Introduction}
Conventional machine learning algorithms generally assume that the training set and the test set are drawn from the same data distribution~\cite{pan2010survey}. The assumption, however, cannot always be guaranteed in real-world applications~\cite{long2015learning,ding2014latent}. To address this, domain adaptation~\cite{pan2011domain,gong2012geodesic,bousmalis2017unsupervised,ding2017robust} has been proposed to mitigate the data distribution shifts among different domains. 

\begin{figure}[t!p]
\begin{center}
\includegraphics[width=0.75\linewidth]{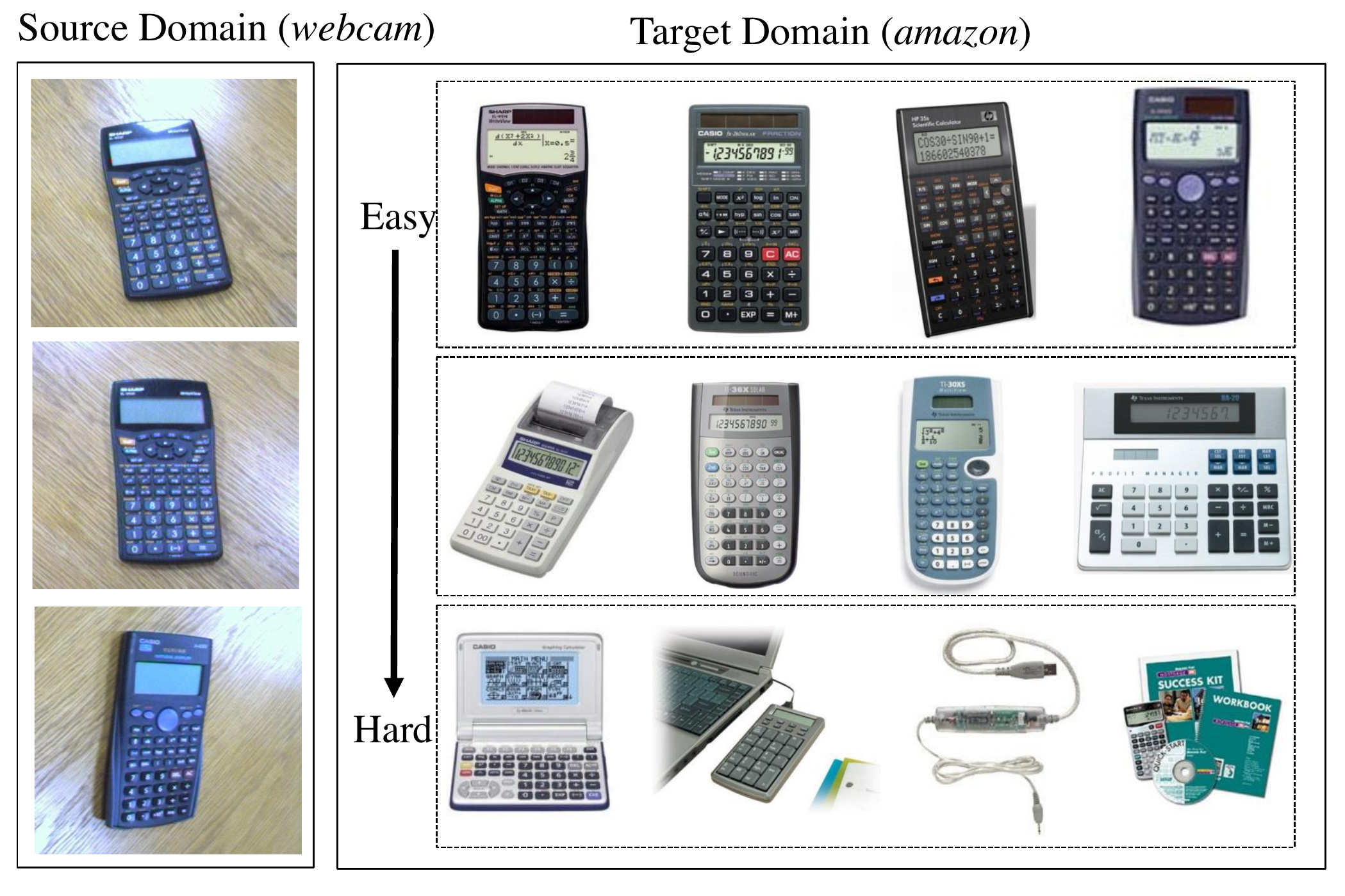}
\end{center}
\vspace{-10pt}
\caption{\small{Illustration of the degrees of classification difficulty. The source domain and target samples are from {\it webcam} dataset and {\it amazon} dataset, respectively. All the shown samples have the same label {\it Calculator}. It is obvious that some target samples are easier to be classified than others when we use the source samples as classification reference.}}
\label{fig:visual}
\vspace{-15pt}
\end{figure}

Existing domain adaptation approaches can be roughly grouped into traditional methods and deep learning methods. Traditional methods~\cite{gong2012geodesic,pan2011domain,li2018transfer} generally do not care how to extract sample features. They only focus on the transfer techniques, such as distribution alignment~\cite{pan2011domain}, feature augmentation~\cite{li2014learning} and landmark selection~\cite{aljundi2015landmarks}. Deep learning methods~\cite{ganin2016domain,ganin2014unsupervised,ding2018semi} take care of both feature extraction and knowledge adaptation via an end-to-end architecture. Yet, no matter what learning paradigm they take, previous domain adaptation methods deploy exactly the same leaning framework for all of the target samples. Specifically, all of the target samples, both easy and hard, are processed via exactly the same pipeline, e.g., the same optimization steps in tradition learning and the same neural network layers in deep learning. Existing methods just feed all the samples into a general formulation which can output an average result. They neglect the inherent degrees of difficulty (as shown in Fig.~\ref{fig:visual}) with the target samples.

In general, a simple or shadow framework is fast but rough. A sophisticated or deep framework, on the contrary, is accurate but slow. It is worth noting that a network shared by both easy and hard samples tends to get deeper and larger since the model has to handle hard samples. For a better understanding, thinking about a smart phone. Although one mostly uses the phone for calls and messages, it has to be powerful enough just in case one wants to play the Temple Run from time to time. As a result, the neglect of different adaptation difficulties makes the domain adaptation methods hard to be deployed in real-time and energy-sensitive applications. Fig.~\ref{fig:visual} has clearly shown that different adaptation difficulties do exist in real-world datasets. Intuitively, if the two domains are highly related, most of the target samples would be easy ones and hard samples tend to be less frequent. Otherwise, one should consider choosing a different source domain for adaptation. Since easy samples can be adapted by a simple or shadow framework and hard samples need more effort, so, why don't we finish the easy ones first with almost no effort and then fully focus on hard ones?



Motivated by the above observations, we propose a novel domain adaptation paradigm which takes the degrees of classification difficulty into consideration. Specifically, we present a deep domain adaptation architecture which has multiple exits. Different exits are located after different layers along the backbone deep network. A common sense about deep neural network is that the features extracted from earlier layers are more general but coarse, while the features extracted from the latter layers are more fine but specific~\cite{long2015learning}. Therefore, the most easy samples are supposed to be classified by very few layers with coarse features, and then these samples can be finished via the first exit. Similarly, the medium hard samples can be handled by the second exit, third exit and so on. At last, the very hard samples are handled by the final exit.

Since deep learning is computing-intensive, more layers generally need more computing power, e.g., GPU, memory and electricity. The early exit paradigm can significantly reduce the computational costs. As a result, it is possible to deploy our solution on a distributed platform. For instance, the first exit can be deployed on edge device, e.g., mobile devices. The second exit on local servers and the final exit on cloud. In particular, we can agilely tailor the deep architecture according to the specific application scenarios. At last, the main contributions of this paper can be listed as follows:

\begin{enumerate}[1)]
 \item We propose a novel learning paradigm for domain adaptation. It explicitly handles the degrees of adaptation difficulty by introducing multiple exits in the learning pipeline. The proposed paradigm can be easily incorporated into deep domain adaptation approaches and significantly reduce their computational costs.  
 \item We present a novel domain adaptation method, i.e., agile domain adaptation networks (ADAN), which puts the learning paradigm into practice. Extensive experiments on open benchmarks verify the effectiveness and efficiency of ADAN. 
 \item For deep transfer learning methods, it is confusing to choose how many layers should be used to extract features. The earlier layers are more transferable but the corresponding features are too coarse. On the contrary, the features extracted from the latter layers are more fine/distinctive but these layers tend to be task-specific and hard to be transfered. In our approach, we find a way out of the dilemma. Specifically, earlier layers are used to classify easy samples and latter layers to classify hard ones. Our formulation takes full advantages from both the early layers and the latter layers. It is a practical way to challenge the fundamental contradiction between the accuracy and speed in domain adaptation tasks.
\end{enumerate}

The remainder of this paper is organized as follows. Section II briefly reviews related work and highlights the merits of our approach. Section III details the proposed learning paradigm and corresponding approach ADAN. Section IV reports the experiments and analyzes ADAN. At last, section~V is the conclusion and future work.

\section{Related Work}

 A typical domain adaptation~\cite{gong2012geodesic,li2018transfer,li2018heterogeneous,li2019locality} problem consists of two domains: a well-labeled source domain and an unlabeled target domain. The two domains generally have the same label space but different data distributions~\cite{pan2010survey}. Domain adaptation aims to mitigate the gap between the two data distributions, so that the knowledge, e.g., features and parameters, learned from the source domain can be transfered to the target domain. Recently, domain adaptation has been successfully applied to many real-world applications, such as image recognition~\cite{hubert2016learning,bousmalis2017unsupervised}, multimedia analysis~\cite{li2014learning,ding2018semi} and recommender systems~\cite{li2017two,li2019zero}.


Since domain adaptation aims to mitigate the distribution gap between the source domain and the target domain, it is vital to find a metric which can measure the data distribution divergence. Maximum mean discrepancy (MMD)~\cite{gretton2012kernel} is widely considered as a favorable criteria in previous work. For instance, deep adaptation networks (DAN)~\cite{long2015learning} generalizes deep convolutional neural networks to the domain adaptation scenario. In DAN, the general (task-invariant) layers are shared by the two domains and the task-specific layers are adapted by multi-kernel MMD. Furthermore, joint adaptation networks (JAN)~\cite{long2017deep} extends DAN by aligning the joint distributions of multiple domain-specific layers across domains based on a joint maximum mean discrepancy (JMMD) criterion.

Recently, generative adversarial networks (GAN)~\cite{goodfellow2014generative} has been introduced into domain adaptation. Compared with the distribution alignment methods, adversarial domain adaptation models~\cite{tzeng2017adversarial,bousmalis2017unsupervised} are able to generate domain invariant features under the supervision of a discriminator. For instance, adversarial discriminative domain adaptation (ADDA)~\cite{tzeng2017adversarial} combines discriminative analysis, untied weight sharing and a GAN loss under a generalized framework. Coupled generative adversarial networks (CoGAN)~\cite{liu2016coupled} minimize the domain shifts by simultaneously training two GANs to handle the source domain and the target domain. 

\begin{figure*}[th]
\begin{center}
\includegraphics[width=0.93\linewidth]{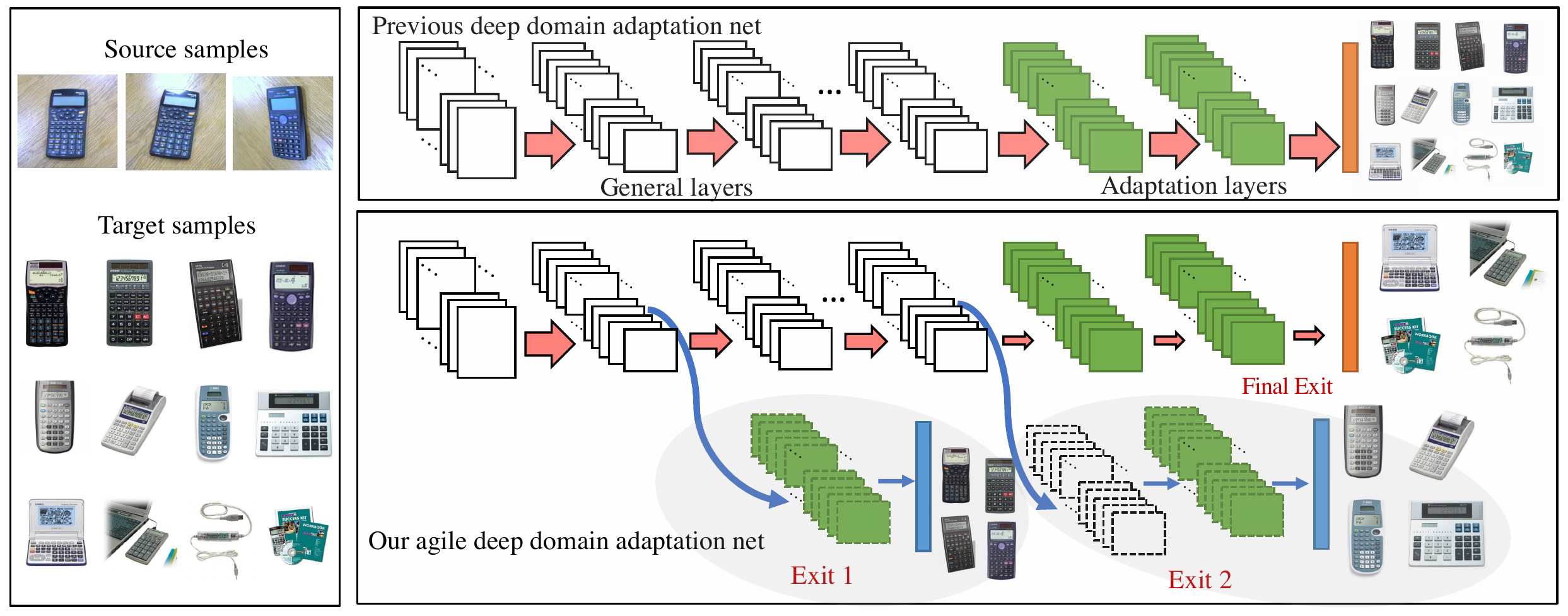}
\end{center}
\vspace{-10pt}
\caption{Main idea of agile domain adaptation networks (ADAN). Please note that this illustration is just for a figurative understanding. We show an ADAN with three exits. The number of exits can be tailored according to specific applications and basic network structures. Different exits share some layers from the backbone network, additional convolutional layers and adaptation layers, as shown by the dash lines, can be added to the exits for fine-tuning. Since some samples exit at the earlier stage in our method, the forward arrows, which also represents computational costs, get thiner and thiner.}
\label{fig:idea}
\vspace{-5pt}
\end{figure*}

No matter MMD loss or GAN loss or both of them are used in domain adaptation networks, every target sample is handled by the same pipeline in previous methods. Different target samples generally have varying visual characteristics even if they are from the same category (as shown in Fig.~\ref{fig:visual}). It is so straightforward that everyone can imagine that some target samples are closer to source samples and some target samples are far away~\cite{gong2013connecting}. Intuitively, the closed samples are easier to be adapted than distant ones. Therefore, we advocate different pipelines for different samples. Specifically, simpler networks for easy and frequent samples, more complex networks for hard and rare ones. From the perspective of network structure, BranchyNet~\cite{teerapittayanon2016branchynet} and hard-aware deeply cascaded embedding (HDC)~\cite{yuan2017hard} are related with our work. BranchyNet leverages the insight that many test samples can be correctly classified early and therefore do not need the later network layers. HDC ensembles a set of models with different complexities in cascaded manner to mine hard examples at multiple levels. However, both BranchyNet and HDC are conventional machine learning approaches. They are limited to scenarios where the training set and the test set are drawn from the same data distribution. In transfer learning tasks, we have to address the distribution gaps along the network layers and exits.


\section{Agile Domain Adaptation}
\subsection{Notations and Definitions}
In this paper, we use a bold uppercase letter and a bold lowercase letter to denote a matrix and a vector, respectively. Subscripts $s$ and $t$ are used to indicate the source domain and the target domain, respectively. We investigate the unsupervised domain adaptation problem defined as follows. 

{\bf Definition 1} {\it A domain $\mathbb{D}$ consists of three parts: feature space $\mathcal{X}$, its probability distribution $P(\mathbf{X})$ and a label set $\mathcal{Y}$, where $\mathbf{X}~\in~\mathcal{X}$. }

 {\bf Problem 1} {\it Given a labeled source domain $\mathbb{D}_s$ and an unlabeled target domain $\mathbb{D}_t$, unsupervised domain adaptation handles the problem of transferring knowledge, e.g., samples, features and parameters, from $\mathbb{D}_s$ to $\mathbb{D}_t$, where $\mathbb{D}_s\!\neq\!\mathbb{D}_t$, $\mathcal{Y}_s\!=\!\mathcal{Y}_t$, $P(\mathbf{X}_{s}) \!\neq\! P(\mathbf{X}_t)$ and $P(\mathbf{y}_s|\mathbf{X}_s)\!\neq\!P(\mathbf{y}_t|\mathbf{X}_t)$.  }


\subsection{Overall Idea}
In unsupervised domain adaptation, we have a labeled source domain $\{\mathbf{X}_s,\mathbf{y}_s\}$ with $n_s$ samples and an unlabeled target domain ${\mathbf{X}_t}$ with $n_t$ samples. Our goal is to train a deep neural network $f(\mathbf{x})$ which has multiple exits, e.g., exit$_1$, exit$_2$, ..., exit$_m$. In the learned deep neural network, the very easy samples can be classified via the first exit, i.e., exit$_1$, which is located after only a few early layers of the backbone network. In the same manner, more difficult samples are handled by subsequent exit$_2$, ..., exit$_{m-1}$ until the last remaining samples are handled by the final exit$_m$. At the testing stage, if samples come in one by one instead of in batches, the sample $\mathbf{x}$ would be first handled by exit$_1$. If exit$_1$ has enough confidence to classify $\mathbf{x}$, the testing would finish. Otherwise, $\mathbf{x}$ will be successively handled by the following exits until one of the exit has high confidence to classify it or finally handled by the last exit. The learning pipeline of this idea is shown in Fig.~\ref{fig:idea}. It is worth noting that the number of exits and the structure of each exit can be tailored according to specific tasks.

\subsection{Problem Formulation}
Since we have the labels of $\mathbf{X}_s$, we can train the deep model on the source domain data in a supervised fashion. Specifically, the empirical error of $f(x)$ on $\mathbf{X}_s$ can be written as:
\begin{equation}
\label{eq:sourceloss}
  \begin{array}{c}
 \mathcal{L}_\mathrm{sup}= \frac{1}{n_s}\sum\limits_{i=1}^{n_s}J(f(\mathbf{x}_{s,i}),\mathbf{y}_{s,i}),
  \end{array} 
\end{equation}
where $J(\cdot)$ is a loss metric. Minimizing $\mathcal{L}_\mathrm{sup}$ can train a model which is suitable for the source domain. However, the model cannot handle the target domain due to the data distribution shift. Therefore, we further introduce domain adaptation layers into the network, so that the trained model can be applied for the target domain. In a deep neural network, e.g., AlexNet~\cite{krizhevsky2012imagenet} and ResNet~\cite{he2016deep}, deep features extracted from the earlier layers are more general and features from the latter layers tend towards specific~\cite{long2015learning}. In other words, the features vary from domain-invariant to domain-specific along the network. Activations in the domain-specific layers are hard to be transferred. Consequently, we remold the domain-specific layers to domain adaptation layers by optimizing a transfer loss $\mathcal{L}_\mathrm{tran}$. As a result, the loss function of each exit$_\epsilon$ $(\epsilon=1,\cdots, m)$ which considers both the source domain and the target domain can be written as:
\begin{equation}
\label{eq:exitloss}
  \begin{array}{c}
 \mathcal{L}_{\mathrm{exit}_\epsilon}= \mathcal{L}_\mathrm{sup} +\lambda \mathcal{L}_\mathrm{tran},
  \end{array} 
\end{equation}
where $\lambda>0$ is the balancing parameter, $\epsilon$ indicates the index of network exits.

In our agile domain adaptation networks, we have multiple exits, e.g., exit$_1$, exit$_2$, ..., exit$_m$. To train the whole ADAN in an end-to-end manner, we jointly optimize the loss functions of each exit. Formally, we optimize a weighted loss sum of $\mathcal{L}_{\mathrm{exit}_\epsilon} (\epsilon=1,2,\cdots,m)$:
\begin{equation}
\label{eq:overallloss}
  \begin{array}{c}
 \mathcal{L}= \sum\limits_{\epsilon=1}^{m}w_\epsilon\mathcal{L}_{\mathrm{exit}_\epsilon},
  \end{array} 
\end{equation}
where $w_\epsilon>0~(\epsilon=1,2,\cdots,m)$ is the loss weight of exit$_\epsilon$. In our paper, we simply set $w_\epsilon =1~(\epsilon=1,2,\cdots,m).$\\

In this paper, we deploy the cross-entropy loss for the labeled source data. If we use ${\mathbf{y}}$ to denote the one-hot ground-truth label vector of sample $\mathbf{x}$, $J(\cdot)$ can be written as:
\begin{equation}
\label{eq:cross-entropy}
  \begin{array}{c}
 J(\hat{\mathbf{y}},{\mathbf{y}})=- \frac{1}{|C|}\sum\limits_{c\in C} y_c \mathrm{log}\hat{y}_c,
  \end{array} 
\end{equation}
where $C$ is the set of all possible labels, $\hat{\mathbf{y}}$ is the predicted label vector of $\mathbf{x}$:
\begin{equation}
\label{eq:softmax}
  \begin{array}{c}
 \hat{\mathbf{y}}= \mathrm{softmax}(f({\mathbf x})) = \dfrac{\mathrm{exp}(f({\mathbf x}))}{\sum\limits_{c\in C} \mathrm{exp}(f({\mathbf x})_c)}.
  \end{array} 
\end{equation}

For the transfer learning part $\mathcal{L}_\mathrm{tran}$, we deploy the multi-kernel MMD~\cite{gretton2012kernel} loss as our metric. Specifically, given two data distributions of the source and the target domain, their MMD can be computed as the square distance between the empirical kernel means as:
\begin{equation}
\label{eq:mmd1}
  \begin{array}{c}
\mathrm{MMD}(\mathbf{X}_s,\mathbf{X}_t) = \frac{1}{n_s^2}\sum\limits_{i=1}^{n_s}\sum\limits_{j=1}^{n_s} k(\mathbf{x}_{s,i},\mathbf{x}_{s,j}) \\ 
 ~~~~~~~~~~~~~~~~~~~~~~	-\frac{2}{n_s n_t}\sum\limits_{i=1}^{n_s}\sum\limits_{j=1}^{n_t} k(\mathbf{x}_{s,i},\mathbf{x}_{t,j}) \\
 ~~~~~~~~~~~~~~~~~~~~~~~~~~	+\frac{1}{n_t^2}\sum\limits_{i=1}^{n_s}\sum\limits_{j=1}^{n_s} k(\mathbf{x}_{t,i},\mathbf{x}_{t,j}),
  \end{array} 
\end{equation}
where $k(\cdot)$ is a kernel function which maps the data features into a reproducing kernel Hilbert space (RKHS). In this paper, we deploy the widely used Gaussian kernel.

\begin{table}[t]
\vspace{8pt}
\centering
\small
\begin{tabular*}{\linewidth}{@{\extracolsep{\fill}}l}
\hline
{\bf Algorithm 1.} {\it Agile Domain Adaptation Networks}\\
\hline
{\bf Training}\\
 ~~~Learning the network parameters by optimizing Eq.~\eqref{eq:overallloss}.\\
{\bf Test}\\
~~~$\epsilon=1$; {\small \it  \hspace{58pt}\%Initialize the network exit index}\\ 
~~~{\it while}  \\
~~~~~~~~$\mathbf{z}$=$f_{\mathrm{exit}_\epsilon}(\mathbf{x})$; {\small \it   \hspace{24pt}\%Get the features from exit$_\epsilon$}\\
~~~~~~~~$\mathbf{y}=\mathrm{softmax}(\mathbf{z})$; {\small \it  ~~~\% Predict the label at exit$_\epsilon$}\\
~~~~~~~~$e=\mathrm{En}(\mathbf{y})$; {\small \it  ~~~~~~~~~~~~\%Calculate the sample entropy}\\ 
~~~~~~~~if $e \le T_\epsilon$ {\small \it  \hspace{36pt}~\%If the entropy is lower than a threshold}\\ 
~~~~~~~~~~~~~{\it return} $\mathbf{y}$; {\small \it  \hspace{25pt}~\%Return the predicted label and finish}\\ 
~~~~~~~~$\epsilon=\epsilon+1$; {\small \it  \hspace{32pt}~\%Otherwise, go to next exit}\\
~~~{\it Until} $\epsilon > m$;\\
~~~{\it return} $\mathbf{y}$;\\
\hline
\end{tabular*}
\label{alg:alm}
\vspace{-5pt}
\end{table}

Eq.~\eqref{eq:mmd1} calculates the MMD on the original data feature $\mathbf{x}$. However, minimizing Eq.~\eqref{eq:mmd1} is implicit in matching the activations generated by the domain adaptation layers. In transfer learning, the source activations and target activations generated by the domain adaptation layers are encouraged to be well-aligned so that the layer parameters can be shared by the two domains. Therefore, we explicitly minimize the MMD on layer activations~\cite{long2017deep}. Let $\mathbf{Z}_s^l$ and $\mathbf{Z}_t^l$ denote the activations generated by layer $l$ from the source domain and the target domain, respectively, the MMD with respect to layer activations can be calculated by:
\begin{equation}
\label{eq:mmd2}
  \begin{array}{c}
 \mathcal{L}_\mathrm{tran}= \frac{1}{n_s^2}\sum\limits_{i=1}^{n_s}\sum\limits_{j=1}^{n_s}\prod\limits_{l=1}^{L} k^l(\mathbf{z}_{s,i}^l,\mathbf{z}_{s,j}^l) \\ 
 ~~~~~~~~	-\frac{2}{n_s n_t}\sum\limits_{i=1}^{n_s}\sum\limits_{j=1}^{n_t}\prod\limits_{l=1}^{L} k^l(\mathbf{z}_{s,i}^l,\mathbf{z}_{t,j}^l) \\
 ~~~~~~~~~~~~	+\frac{1}{n_t^2}\sum\limits_{i=1}^{n_s}\sum\limits_{j=1}^{n_s}\prod\limits_{l=1}^{L} k^l(\mathbf{z}_{t,i}^l,\mathbf{z}_{t,j}^l).
  \end{array} 
\end{equation}

Furthermore, Eq.~\eqref{eq:mmd2} can be rewritten as the following equivalent form to reduce the computational costs~\cite{gretton2012kernel}. 
\begin{equation}
\label{eq:mmd3}
  \begin{array}{c}
 \mathcal{L}_\mathrm{tran}= \\~~\frac{2}{n_s}\sum\limits_{i=1}^{n_s/2}\Big(\prod\limits_{l=1}^{L} k^l(\mathbf{z}_{s,2i-1}^l,\mathbf{z}_{s,2i}^l) + \prod\limits_{l=1}^{L} k^l(\mathbf{z}_{t,2i-1}^l,\mathbf{z}_{t,2i}^l)\Big) \\ 
 -\frac{2}{n_s}\sum\limits_{i=1}^{n_s/2}\Big(\prod\limits_{l=1}^{L} k^l(\mathbf{z}_{s,2i-1}^l,\mathbf{z}_{t,2i}^l) + \prod\limits_{l=1}^{L} k^l(\mathbf{z}_{t,2i-1}^l,\mathbf{z}_{s,2i}^l)\Big) .
\end{array} 
\end{equation}

With Eq.~\eqref{eq:sourceloss} and Eq.~\eqref{eq:mmd3}, we can train our agile domain adaptation networks by optimizing the overall loss as shown in Eq.~\eqref{eq:overallloss}.\\ 

 For the earlier detections, we need to estimate whether a sample should be finished at each exit. Therefore, we need a metric to measure the classification confidence. In this paper, we use the sample entropy as the metric, which is defined as:
\begin{equation}
\label{eq:sample-entropy}
  \begin{array}{c}
 \mathrm{En}({\mathbf{y}})= -\sum\limits_{c\in C} y_c \mathrm{log}y_c,
  \end{array} 
\end{equation}
where $\mathbf{y}$ is a label vector which consists of the probabilities of all possible labels computed at each exit. From the physical meaning of entropy, we know that a lower entropy indicts a more certain output. As a result, we compare the sample entropy with a threshold in each exit. If the entropy is lower than the threshold, the classification of the sample would be finished. Otherwise, the sample will goto the next exit for prediction. For a better understanding, we show the main steps of our ADAN in Algorithm~1.

\section{Experiments}
In this section, we verify the proposed method with both accuracy and efficiency results. Two widely used base architectures, e.g., the classical LeNet~\cite{lecun1998gradient} and the state-of-the-art ResNet~\cite{he2016deep}, are tailored to work in the manner of agile domain adaptation. The source codes will be publicly available on our GitHub page.

\subsection{Data Description}
{\bf USPS} and {\bf MNIST} are two widely used datasets in domain adaptation. Both of them are comprised of images of handwritten digits. There are 9,298 and 70,000 samples in total in each dataset, respectively.

\noindent{\bf Office-31} dataset~\cite{saenko2010adapting} consists of 4,652 samples from 31 categories. Samples in this dataset are from 3 subsets, e.g., Amazon ({\bf A}), DSLR ({\bf D}) and Webcam ({\bf W}). Specifically, Amazon includes images downloaded from amazon.com. DSLR consists of samples captured by a digital SLR camera. Images in Webcam are shoot by a  low-resolution web camera.

\subsection{Implementation Details}
Our proposed paradigm is independent from specific base architectures. It can be easily incorporated into any popular deep networks. Limited by space, we implement our ADAN based on two base architectures, e.g., LeNet~\cite{lecun1998gradient} for digits recognition and ResNet~\cite{he2016deep} for object classification.

LeNet-5 contains three convolutional layers and two fully connected layers. In our ADAN, we add one earlier exit after the first convolutional layer. The earlier exit consists of a convolutional layer and a fully connected layer. For LeNet-5, we set batchsize as $128$, learning rate as $0.001$, optimizer as SGD with momentum $0.9$ and weight decay $0.0001$.

For ResNet-50, we add two earlier exits into the backbone network. The first one is located after the 3rd layer, and the second one after the 39th layer. One can also tailor the earlier exits, either numbers or locations, for their own applications. In our implementation, we provide a general template, which makes it embarrassingly simple to add, remove and modify the earlier exits. With respect to the training parameters, we set batchsize to $24$. We also use SGD with momentum of $0.9$ to optimize the objective function. The learning rate is adjusted during SGD using the same formula as reported in~\cite{long2017deep}.

\subsection{Metrics and Compared Methods}
In the domain adaptation community, accuracy of the target domain is the most widely used metric. For the sake of fairness, we follow previous work~\cite{tzeng2017adversarial,long2017deep} and also report the accuracy of the target domain. Each of the reported results of our method is an average of $5$ runs. The results of compared method are cited from the original paper. Or, if not available in the original paper, we report the best results we can achieve by running their codes. All of the experiments settings are following the previous work~\cite{bousmalis2017unsupervised,long2017deep} to ensure fair comparison.

The following state-of-the-art methods are reported for comparison:
\begin{itemize}
	\item TCA~\cite{pan2011domain} and GFK~\cite{gong2012geodesic} are two representative traditional domain adaptation methods. For fair comparison, deep features extracted from ResNet-50 are used as their raw inputs.
	\item LeNet-5~\cite{lecun1998gradient} and ResNet~\cite{he2016deep} are two baselines. BranchyNet~\cite{teerapittayanon2016branchynet} is a non-transfer method which deploys the similar idea with ours. By reporting BranchyNet as a baseline, we can see that transfer learning tasks need special care.
	\item CORAL~\cite{sun2016return}, DANN~\cite{ganin2014unsupervised}, DDC~\cite{tzeng2014deep}, DAN~\cite{long2015learning} and JAN~\cite{long2017deep} are state-of-the-art deep domain adaptation approaches.
\end{itemize}

\renewcommand{\arraystretch}{0.9}
\begin{table}[t!p]
\centering
\caption{Results (accuracy \%) of digits recognition on MNIST and USPS datasets with LeNet-5.}
\vspace{-5pt}
\label{tab:digits1}
\footnotesize\begin{tabular}{|c|c|}
\hline
Method & MNIST$\rightarrow$USPS \\
\hline \hline
Source Only & 78.9 \\
\hline
BranchyNet~(Teerapittayanon et al. 2106) & 79.6 \\
\hline
DAN~\cite{long2015learning} & 81.1 \\
\hline
CORAL~\cite{sun2016return} & 81.7 \\
\hline 
DANN~\cite{ganin2016domain}  & 85.1    \\
\hline
{\bf ADAN (Ours)}  &  {\bf 91.3} \\
\hline
\end{tabular}
\end{table}

\begin{table}[t!p]
\centering
\caption{Our results of MNIST$\rightarrow$USPS with different thresholds, in which {\it Acc.} is the overall accuracy, {\it time} is the running time and {\it ratio} is the ratio of exited target samples via the first earlier exit. }
\vspace{-5pt}
\label{tab:digits2}
\footnotesize\begin{tabular}{|c|c|c|c|c|c|c|}
\hline
Threshold & Acc.(\%) & Time (ms) & Ratio (\%) & Speedup  \\
\hline \hline
-Baseline- & 91.27 & 2.42 & 0 & 1x \\
\hline
0.005 & 91.56 & 2.08 & 16.09 & 1.16x\\
\hline
0.010 & 91.56 & 1.91 & 19.02 & 1.27x \\
\hline
0.025 & 91.57 & 1.33 & 34.81 & 1.81x \\
\hline
0.050 & 91.60 & 1.14 & 47.27 & 2.12x \\
\hline
0.100 & 91.62 & 1.03 & 57.35 & 2.35x \\
\hline
0.250 & {\bf 91.62} & 0.83 & 69.31 & {\bf 2.92x} \\
\hline
0.750 & 90.06 & 0.74 & 86.12 & 3.29x \\
\hline
1.500 & 87.55 & 0.44 & 99.50 & 5.52x \\
\hline
\end{tabular}
\vspace{-10pt}
\end{table}

\begin{table*}[ht!p]
\centering
\caption{Results (accuracy \%) of object classification on Office-31 dataset with ResNet.}
\vspace{-7pt}
\label{tab:office}
\footnotesize\begin{tabular}{|c|c|c|c|c|c|c|c|}
\hline
Method & A$\rightarrow$D & A$\rightarrow$W & D$\rightarrow$A & D$\rightarrow$W & W$\rightarrow$A & W$\rightarrow$D & Average \\
\hline \hline
ResNet & $68.9 \pm 0.2$ & $ 68.4 \pm 0.2$& $62.5 \pm 0.3$ & $96.7 \pm 0.1$ & $60.7 \pm 0.3$ & $99.3 \pm 0.1$ & $76.1 \pm 0.2$\\ 
\hline
TCA   & $74.1 \pm 0.0$ & $72.7 \pm 0.0$ & $61.7 \pm 0.0$ & $96.7 \pm 0.0$ & $60.9 \pm 0.0$ & $99.6 \pm 0.0$ & $77.6 \pm 0.0$ \\
\hline
GFK & ~$74.5 \pm 0.0$~ & ~$72.8 \pm 0.0$~ & ~$63.4 \pm 0.0$~ & ~$95.0 \pm 0.0$~ & ~$61.0 \pm 0.0$~ & ~$98.2 \pm 0.0$~  & $77.5 \pm 0.0$ \\  
\hline
DDC & $76.5 \pm 0.3$ & $75.6 \pm 0.2$ & $62.2 \pm 0.4$ & $96.0 \pm 0.2$ & $61.5 \pm 0.5$ & $98.2 \pm 0.1$ & $78.3 \pm 0.3$ \\
\hline
DAN & $78.6 \pm 0.2$ & $80.5 \pm 0.4$ & $63.6 \pm 0.3$ & $97.1 \pm 0.2$ & $62.8 \pm 0.2$ & $99.6 \pm 0.1$ & $80.4 \pm 0.2$ \\
\hline
~~~~~RevGrad~~~~~ & ~~$79.7 \pm 0.4$~~ & ~~$82.0 \pm 0.4$~~ & ~~$68.2 \pm 0.4$~~ & ~~$96.9 \pm 0.2$~~ & ~~$67.4 \pm 0.5$~~ & ~~$99.1 \pm 0.1$~~ & ~~$82.2 \pm 0.3$~~\\ 
\hline
JAN & $84.7 \pm 0.3$ & $85.4 \pm 0.3$& $68.6 \pm 0.3$ & ${\bf 97.4 \pm 0.2}$ & $70.0 \pm 0.4$ & ${\bf 99.8 \pm 0.2}$ & $84.3 \pm 0.2$\\ 
\hline
{\bf Ours}  & ${\bf 85.8 \pm 0.3}$ & ${\bf 88.8 \pm 0.4}$ & ${\bf 69.8 \pm 0.3}$ & ${97.4 \pm 0.2}$ & ${\bf 70.6 \pm 0.3}$ & ${99.4 \pm 0.4}$ & ${\bf 85.3 \pm 0.3}$\\
\hline
\end{tabular}
\vspace{-5pt}
\end{table*}

\begin{figure*}[th]
\begin{center}
\subfigure{
\includegraphics[width=0.98\linewidth]{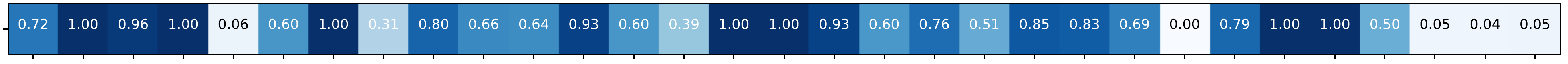}
}
\subfigure{
\vspace{-15pt}
\includegraphics[width=0.98\linewidth]{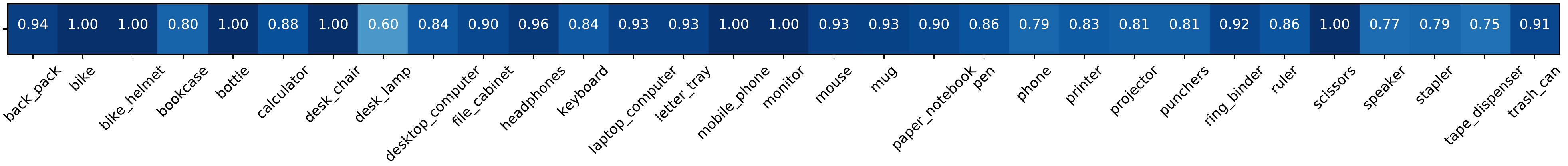}
}
\end{center}
\vspace{-20pt}
\caption{The class-wise accuracy on A$\rightarrow$W. Darker color denotes higher accuracy. The first line shows the results of ResNet, and the last line shows the results of our ADAN. }
\label{fig:visualization}
\vspace{-10pt}
\end{figure*}

\begin{table}[ht!p]
\centering
\caption{Our results of object classification on Office-31 dataset with different thresholds, in which {\it threshold} {a/b} denotes a\% and b\% samples are exited via the first earlier exit and the second earlier exit, respectively. The remainder, i.e., (100-a-b)\%, is processed by the final exit. }
\vspace{-5pt}
\label{tab:office2}
\footnotesize\begin{tabular}{|c|c|c|c|c|c|}
\hline
Evaluation &  \multicolumn{2}{|c|}{W$\rightarrow$D} &  \multicolumn{2}{|c|}{D$\rightarrow$W} \\
\hline
Threshold & ~~Acc.(\%)~~ & Speedup & ~~Acc.(\%)~~  & Speedup  \\
\hline \hline
-Baseline- & 99.39 & 1.0x & 97.36 & 1.0x \\
\hline
10/50 & 98.74 & 1.73x & 96.69 & 1.32x \\
\hline
10/60 & 98.54 & 1.82x & 96.63 & 1.31x \\
\hline
10/70 & 98.59 & 2.17x & 96.32 & 1.36x \\
\hline
20/50 & 97.99 & 1.63x & 94.27 & 1.34x\\
\hline 
20/60 & 97.23 & 2.52x & 94.49 & 1.35x \\
\hline
30/40 & 96.33 & 2.03x & 91.88 & 1.36x \\
\hline
30/50 & 94.72 & 1.99x & 90.47 & 1.22x \\
\hline
40/50 & 91.86 & 1.71x & 89.05 & 1.24x \\
\hline
50/20 & 88.95 & 1.74x & 85.60 & 1.21x \\
\hline
50/30 & 87.04 & 2.37x & 84.43 & 1.20x \\
\hline
\end{tabular}
\vspace{-15pt}
\end{table}

\subsection{Quantitative Results}

Table~\ref{tab:digits1} and Table~\ref{tab:digits2} report the experimental results of our method on MNIST and USPS datasets. Specifically, Table~\ref{tab:digits1} verifies that our approach can achieve state-of-the-art accuracy. Table~\ref{tab:digits2} further demonstrates that we can speed up the model THREE TIMES while the accuracy does not drop.

From the results in Table~\ref{tab:digits2}, several interesting observations can be drawn. At first, both the accuracy and the speed are improved by adding earlier exits. The backbone network achieves 91.27\% in 2.42ms. With a earlier exit, however, ADAN can achieve 91.62\% in only 0.83ms. In this case, 69.31\% samples are finished by the earlier exit. Secondly, if we set a relatively big entropy threshold, e.g., 1.5, 99.5\% samples can be handled by the earlier exit with 5.52 times speedup, while the accuracy only drops 3.72\%. In real-world applications, we can deploy the earlier exit on mobile device and handle most of the cases with sufficient accuracy. At the same time, the complete backbone network can be deployed on the cloud to take care of hard samples. The last but not the least, since the features vary from domain-invariant to domain-specific along a deep network~\cite{long2015learning}, features extracted from either earlier layers or latter layers have their cons and pros in transfer learning tasks. Our approach, notably, can take advantage from every layer and maximize the value of deep networks.

Table~\ref{tab:office} and Table~\ref{tab:office2} show the results on Office-31 dataset. Similar to the results of digits recognition, we can observe that our approach achieves state-of-the-art performance and it can significantly reduce the running time by the earlier exits. Compared with the MNIST and USPS dataset, Office-31 is much more challenging. As a result, if the ratio of earlier exited samples goes high, the accuracy dropping is more obvious than in the digits recognition task. Notice that the baseline (backbone network) has 50 layers, while the first exit in our model is located after the 3rd layer, which is very shallow compared with ResNet-50. However, our model still get the accuracy of 97.23\% compared with the baseline 99.8\% on W$\rightarrow$D when we set 20\% and 60\% samples exit from the first earlier exit and the second earlier exit, respectively. Notably, with the sacrifice of 1.6\% accuracy, we can speed up the model 2.52 times! The results on D$\rightarrow$W and other evaluations, e.g., A$\rightarrow$W and A$\rightarrow$D, draw the same conclusion. It is worth noting that {\it Amazon} is a very challenging dataset, the results on A$\rightarrow$W and A$\rightarrow$D, therefore, are not outstanding as the results on {\it DSLR} and {\it Webcam}. In our experiments, we tried that if we move the first exits backward several layers or if reduce the ratio of earlier detection, the results on A$\rightarrow$W and A$\rightarrow$D would improve. In our source code, we provide a general template to modify, add and remove earlier exits. One can tailor a personal ADAN with few lines of python codes. 

\begin{figure*}[th]
\begin{center}
\includegraphics[width=0.9\linewidth]{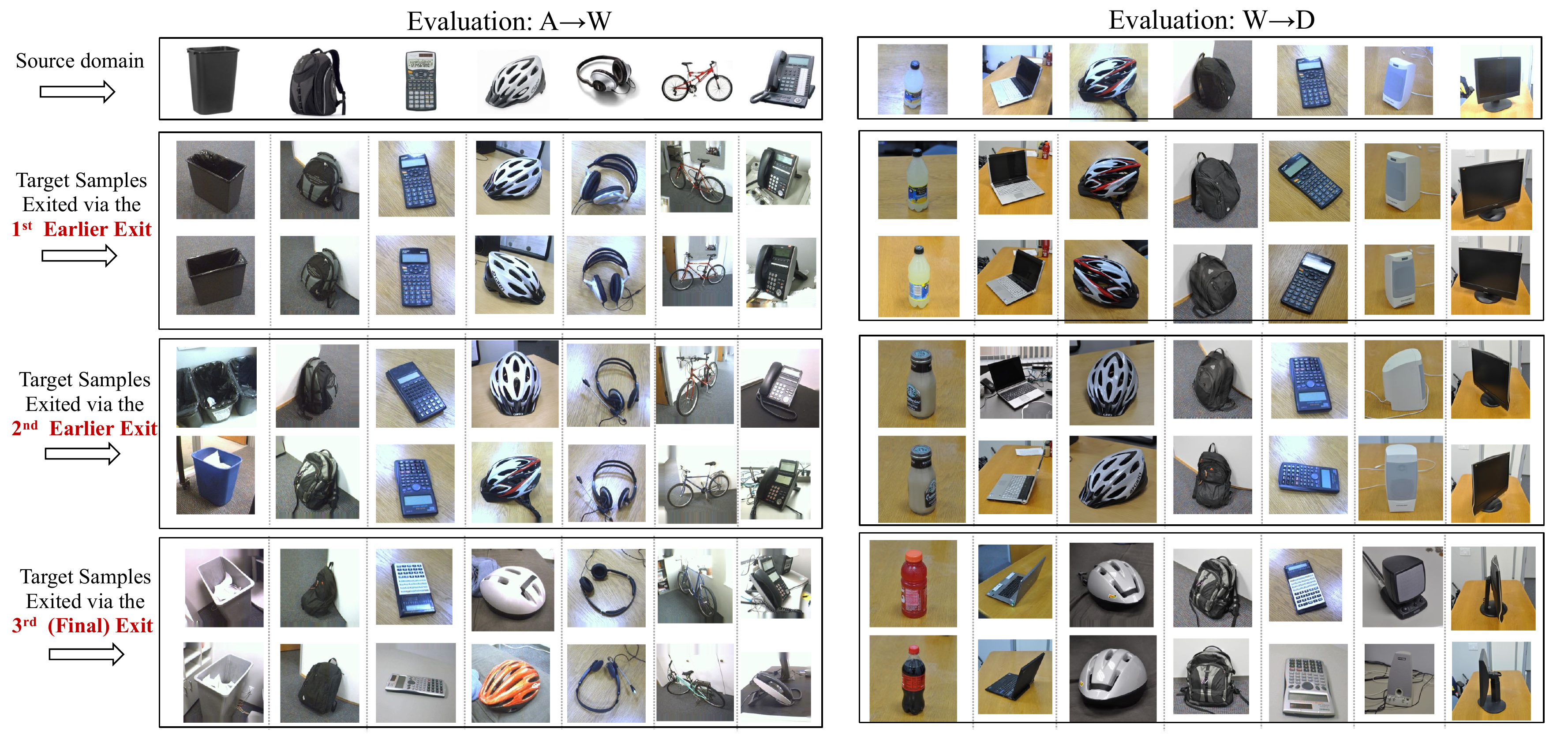}
\end{center}
\vspace{-10pt}
\caption{Representative images of the source domain samples, and the target domain samples which are exited via the 1st, 2nd earlier exit and the final exit.}
\label{fig:samples}
\vspace{-10pt}
\end{figure*}

\begin{figure}[th]
\begin{center}
\subfigure[Parameter Sensitivity]{
\includegraphics[width=0.75\linewidth]{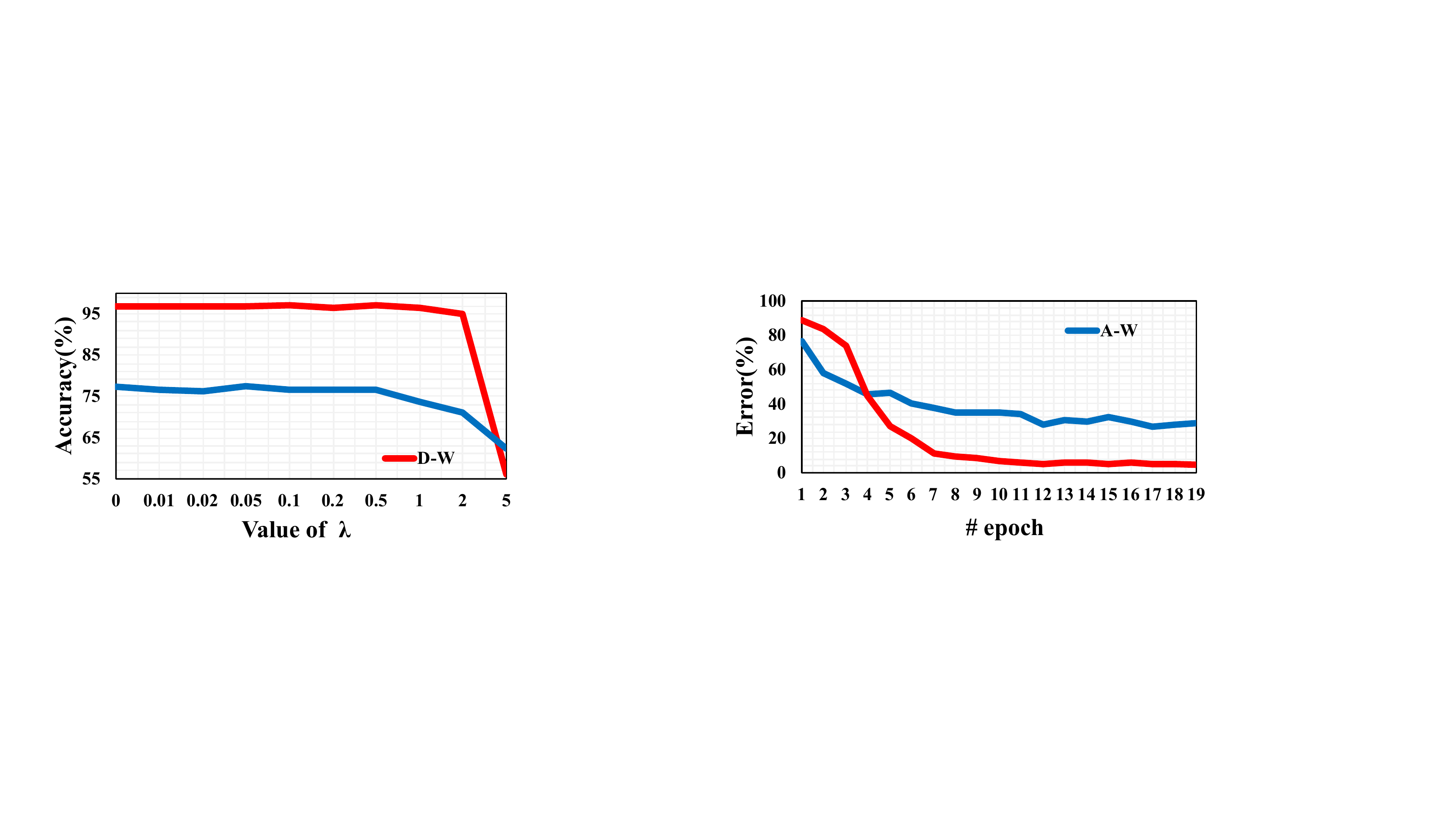}
}
\subfigure[Convergence Curve]{
\vspace{-15pt}
\includegraphics[width=0.73\linewidth]{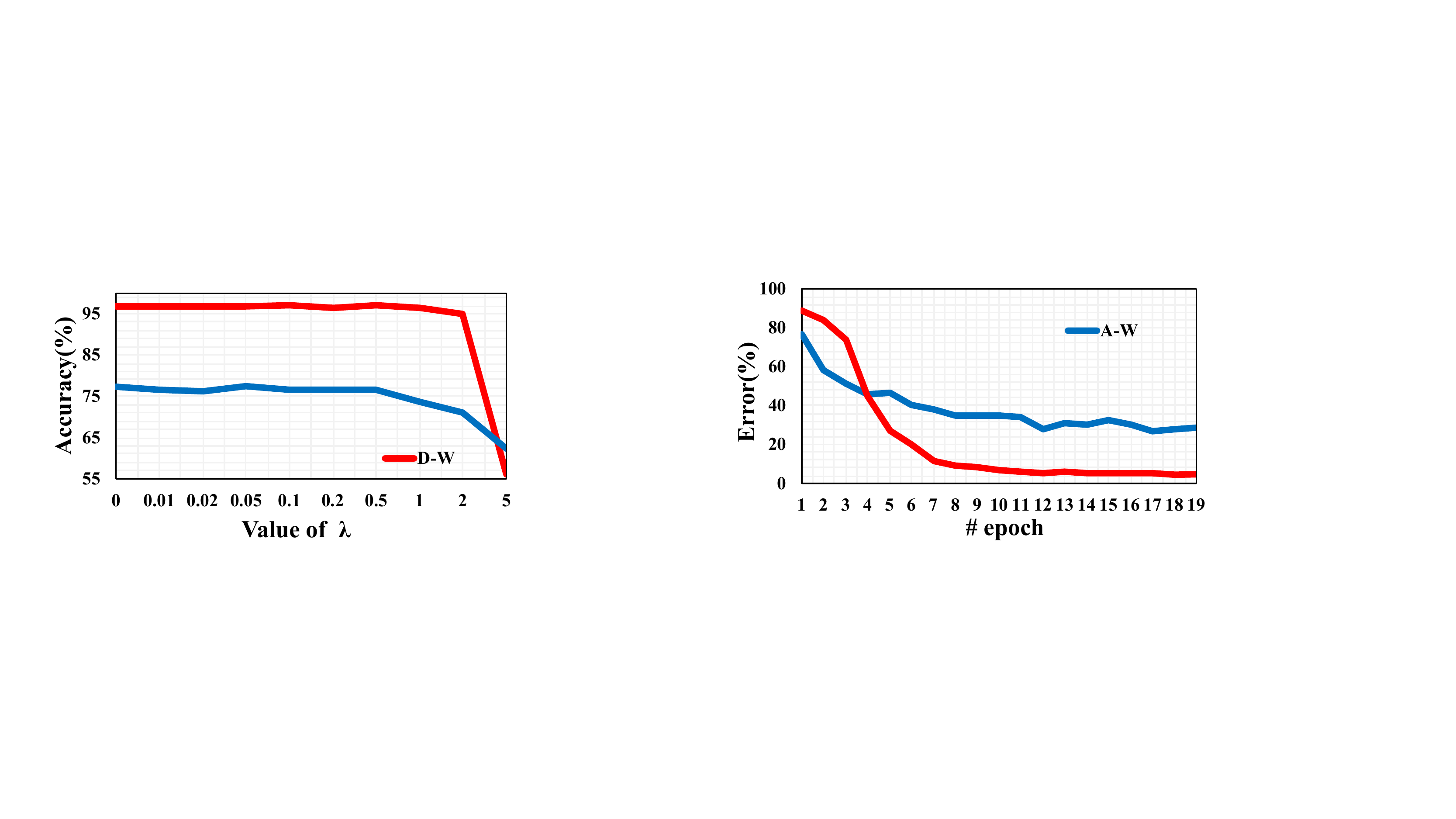}
}
\end{center}
\vspace{-15pt}
\caption{Parameter sensitivity and convergence curve.}
\label{fig:para}
\vspace{-15pt}
\end{figure}

\subsection{Qualitative Results}
{\bf Back to the Samples.} The very basic motivation behind our formulation is that our approach handles the degrees of adaptation difficulty by introducing multiple earlier exits into the learning pipeline. To verify that our approach does perceive the adaptation difficulty and handle it with different strategies, we show some samples exited from different earlier exits in Fig.~\ref{fig:samples}. From the results, it is crystal clear that our approach is able to identify the degrees of adaptation difficulty. For instance, let us take the last column, i.e., samples with the label {\it monitor}, on evaluation W$\rightarrow$D as an example. We can see that the samples exited via the first earlier exit are very similar with the source samples. These samples are easy to be adapted. The target samples exited from the second earlier exit are slightly different from the source samples in view point. At last, the samples exited from the final exit have distinctive capture angles with the source samples. These target samples are hard to be adapted. For the easy samples, we only use few layers to speed up the model. At the same time, for the hard samples, we deploy more deep networks to guarantee the accuracy. Our approach is agile and resilient.

{\bf Class-wise Visualization.} Fig.~\ref{fig:visualization} reports the visualized class-wise accuracy. Comparing the two lines of Fig.~\ref{fig:visualization}, it is clear that our approach is able to mitigate the distribution gaps between the source domain and the target domain.

{\bf Parameter Sensitivity.} The main parameter in our model is $\lambda$ which balances the weight of supervised loss and transfer loss. Fig.~\ref{fig:para}(a) reports the effects of different $\lambda$ values. It can be seen that our model is not sensitive to the parameter when it chosen from [0,1]. However, the performance will degrade when $\lambda>2$. Since the model is targeted at leveraging knowledge from the source domain to facilitate the target domain, a large $\lambda$ would weaken the contribution of the source domain. The other hyper-parameter in our formulation is the weight $w_i$, we simply set it to one throughout this paper. We also tested it for different settings, and find $w_i$ is not sensitive to the overall performance.

{\bf Convergence.} Fig.~\ref{fig:para}(a) reports the overall error with respect to different epochs. The results reflect the convergence of our approach. From the results we can observe that our approach is able to achieve the stable result around 20 epochs. It is worth noting that the reported results are from the evaluations with 10\% and 50\% samples exited via the first and second earlier exit, respectively. Thus, the results are slightly worse than the backbone network. However, the results report the overall convergence rather than only the backbone.

{\bf Ablation Analysis.} Our approach consists of the backbone networks and the earlier exits. If we remove the exits and only remain the backbone, the performance of our approach would be similar with results reported in Table~\ref{tab:office} and the baseline in Table~\ref{tab:office2}. However, the running cost would be an issue for mobile devices and real-time systems.

\section{Conclusion}
In this paper, we propose a novel learning paradigm for domain adaptation. In the proposed paradigm, several earlier detections are performed to identify the adaptation difficulty. At the same time, several earlier exit are added along the backbone network to reduce the running cost. The proposed paradigm can be easily incorporated into existing deep domain adaptation networks or even non-transfer deep architectures, e.g., LeNet, AlexNet and ResNet.

In addition, a novel domain adaptation approach, i.e., agile domain adaptation networks (ADAN), is implemented to verify the effectiveness and efficiency of the proposed paradigm. Extensive experiment results, both quantitative and qualitative, show that our approach is able to achieve state-of-the-art accuracy and significantly reduce the running cost at the same time. In our future work, we will investigate the self-adapted ADAN which can automatically and dynamically adjust the earlier exits.

\balance
\bibliography{aaai19}

\begin{thebibliography}{}

\bibitem[\protect\citeauthoryear{Aljundi \bgroup et al\mbox.\egroup
  }{2015}]{aljundi2015landmarks}
Aljundi, R.; Emonet, R.; Muselet, D.; and Sebban, M.
\newblock 2015.
\newblock Landmarks-based kernelized subspace alignment for unsupervised domain
  adaptation.
\newblock In {\em CVPR},  56--63.

\bibitem[\protect\citeauthoryear{Bousmalis \bgroup et al\mbox.\egroup
  }{2017}]{bousmalis2017unsupervised}
Bousmalis, K.; Silberman, N.; Dohan, D.; Erhan, D.; and Krishnan, D.
\newblock 2017.
\newblock Unsupervised pixel-level domain adaptation with generative
  adversarial networks.
\newblock In {\em CVPR}, volume~1,  7--16.

\bibitem[\protect\citeauthoryear{Ding and Fu}{2017}]{ding2017robust}
Ding, Z., and Fu, Y.
\newblock 2017.
\newblock Robust transfer metric learning for image classification.
\newblock {\em IEEE TIP} 26(2):660--670.

\bibitem[\protect\citeauthoryear{Ding, Nasrabadi, and Fu}{2018}]{ding2018semi}
Ding, Z.; Nasrabadi, N.~M.; and Fu, Y.
\newblock 2018.
\newblock Semi-supervised deep domain adaptation via coupled neural networks.
\newblock {\em IEEE TIP} 27(11):5214--5224.

\bibitem[\protect\citeauthoryear{Ding, Shao, and Fu}{2014}]{ding2014latent}
Ding, Z.; Shao, M.; and Fu, Y.
\newblock 2014.
\newblock Latent low-rank transfer subspace learning for missing modality
  recognition.
\newblock In {\em AAAI},  1192--1198.

\bibitem[\protect\citeauthoryear{Ganin and
  Lempitsky}{2014}]{ganin2014unsupervised}
Ganin, Y., and Lempitsky, V.
\newblock 2014.
\newblock Unsupervised domain adaptation by backpropagation.
\newblock {\em arXiv preprint arXiv:1409.7495}.

\bibitem[\protect\citeauthoryear{Ganin \bgroup et al\mbox.\egroup
  }{2016}]{ganin2016domain}
Ganin, Y.; Ustinova, E.; Ajakan, H.; Germain, P.; Larochelle, H.; Laviolette,
  F.; Marchand, M.; and Lempitsky, V.
\newblock 2016.
\newblock Domain-adversarial training of neural networks.
\newblock {\em JMLR} 17(1):2096--2030.

\bibitem[\protect\citeauthoryear{Gong \bgroup et al\mbox.\egroup
  }{2012}]{gong2012geodesic}
Gong, B.; Shi, Y.; Sha, F.; and Grauman, K.
\newblock 2012.
\newblock Geodesic flow kernel for unsupervised domain adaptation.
\newblock In {\em CVPR},  2066--2073.
\newblock IEEE.

\bibitem[\protect\citeauthoryear{Gong, Grauman, and
  Sha}{2013}]{gong2013connecting}
Gong, B.; Grauman, K.; and Sha, F.
\newblock 2013.
\newblock Connecting the dots with landmarks: Discriminatively learning
  domain-invariant features for unsupervised domain adaptation.
\newblock In {\em ICML},  222--230.

\bibitem[\protect\citeauthoryear{Goodfellow \bgroup et al\mbox.\egroup
  }{2014}]{goodfellow2014generative}
Goodfellow, I.; Pouget-Abadie, J.; Mirza, M.; Xu, B.; Warde-Farley, D.; Ozair,
  S.; Courville, A.; and Bengio, Y.
\newblock 2014.
\newblock Generative adversarial nets.
\newblock In {\em NIPS},  2672--2680.

\bibitem[\protect\citeauthoryear{Gretton \bgroup et al\mbox.\egroup
  }{2012}]{gretton2012kernel}
Gretton, A.; Borgwardt, K.~M.; Rasch, M.~J.; Sch{\"o}lkopf, B.; and Smola, A.
\newblock 2012.
\newblock A kernel two-sample test.
\newblock {\em JMLR} 13(Mar):723--773.

\bibitem[\protect\citeauthoryear{He \bgroup et al\mbox.\egroup
  }{2016}]{he2016deep}
He, K.; Zhang, X.; Ren, S.; and Sun, J.
\newblock 2016.
\newblock Deep residual learning for image recognition.
\newblock In {\em CVPR},  770--778.

\bibitem[\protect\citeauthoryear{Hubert~Tsai, Yeh, and
  Frank~Wang}{2016}]{hubert2016learning}
Hubert~Tsai, Y.-H.; Yeh, Y.-R.; and Frank~Wang, Y.-C.
\newblock 2016.
\newblock Learning cross-domain landmarks for heterogeneous domain adaptation.
\newblock In {\em CVPR},  5081--5090.

\bibitem[\protect\citeauthoryear{Krizhevsky, Sutskever, and
  Hinton}{2012}]{krizhevsky2012imagenet}
Krizhevsky, A.; Sutskever, I.; and Hinton, G.~E.
\newblock 2012.
\newblock Imagenet classification with deep convolutional neural networks.
\newblock In {\em NIPS},  1097--1105.

\bibitem[\protect\citeauthoryear{LeCun \bgroup et al\mbox.\egroup
  }{1998}]{lecun1998gradient}
LeCun, Y.; Bottou, L.; Bengio, Y.; and Haffner, P.
\newblock 1998.
\newblock Gradient-based learning applied to document recognition.
\newblock {\em Proceedings of the IEEE} 86(11):2278--2324.

\bibitem[\protect\citeauthoryear{Li \bgroup et al\mbox.\egroup
  }{2014}]{li2014learning}
Li, W.; Duan, L.; Xu, D.; and Tsang, I.~W.
\newblock 2014.
\newblock Learning with augmented features for supervised and semi-supervised
  heterogeneous domain adaptation.
\newblock {\em TPAMI} 36(6):1134--1148.

\bibitem[\protect\citeauthoryear{Li \bgroup et al\mbox.\egroup
  }{2017}]{li2017two}
Li, J.; Lu, K.; Huang, Z.; and Shen, H.~T.
\newblock 2017.
\newblock Two birds one stone: on both cold-start and long-tail recommendation.
\newblock In {\em ACM MM},  898--906.
\newblock ACM.

\bibitem[\protect\citeauthoryear{Li \bgroup et al\mbox.\egroup
  }{2018a}]{li2018heterogeneous}
Li, J.; Lu, K.; Huang, Z.; Zhu, L.; and Shen, H.~T.
\newblock 2018a.
\newblock Heterogeneous domain adaptation through progressive alignment.
\newblock {\em IEEE TNNLS} 30(5):1381--1391.

\bibitem[\protect\citeauthoryear{Li \bgroup et al\mbox.\egroup
  }{2018b}]{li2018transfer}
Li, J.; Lu, K.; Huang, Z.; Zhu, L.; and Shen, H.~T.
\newblock 2018b.
\newblock Transfer independently together: A generalized framework for domain
  adaptation.
\newblock {\em IEEE TCYB}.

\bibitem[\protect\citeauthoryear{Li \bgroup et al\mbox.\egroup
  }{2019a}]{li2019locality}
Li, J.; Jing, M.; Lu, K.; Zhu, L.; and Shen, H.~T.
\newblock 2019a.
\newblock Locality preserving joint transfer for domain adaptation.
\newblock {\em IEEE TIP}.

\bibitem[\protect\citeauthoryear{Li \bgroup et al\mbox.\egroup
  }{2019b}]{li2019zero}
Li, J.; Jing, M.; Lu, K.; Zhu, L.; Yang, Y.; and Huang, Z.
\newblock 2019b.
\newblock From zero-shot learning to cold-start recommendation.
\newblock {\em AAAI}.

\bibitem[\protect\citeauthoryear{Liu and Tuzel}{2016}]{liu2016coupled}
Liu, M.-Y., and Tuzel, O.
\newblock 2016.
\newblock Coupled generative adversarial networks.
\newblock In {\em NIPS},  469--477.

\bibitem[\protect\citeauthoryear{Long \bgroup et al\mbox.\egroup
  }{2015}]{long2015learning}
Long, M.; Cao, Y.; Wang, J.; and Jordan, M.~I.
\newblock 2015.
\newblock Learning transferable features with deep adaptation networks.
\newblock In {\em ICML},  97--105.

\bibitem[\protect\citeauthoryear{Long \bgroup et al\mbox.\egroup
  }{2017}]{long2017deep}
Long, M.; Zhu, H.; Wang, J.; and Jordan, M.~I.
\newblock 2017.
\newblock Deep transfer learning with joint adaptation networks.
\newblock In {\em ICML},  2208--2217.

\bibitem[\protect\citeauthoryear{Pan \bgroup et al\mbox.\egroup
  }{2011}]{pan2011domain}
Pan, S.~J.; Tsang, I.~W.; Kwok, J.~T.; and Yang, Q.
\newblock 2011.
\newblock Domain adaptation via transfer component analysis.
\newblock {\em IEEE TNN} 22(2):199--210.

\bibitem[\protect\citeauthoryear{Pan, Yang, and others}{2010}]{pan2010survey}
Pan, S.~J.; Yang, Q.; et~al.
\newblock 2010.
\newblock A survey on transfer learning.
\newblock {\em IEEE TKDE} 22(10):1345--1359.

\bibitem[\protect\citeauthoryear{Saenko \bgroup et al\mbox.\egroup
  }{2010}]{saenko2010adapting}
Saenko, K.; Kulis, B.; Fritz, M.; and Darrell, T.
\newblock 2010.
\newblock Adapting visual category models to new domains.
\newblock In {\em ECCV},  213--226.
\newblock Springer.

\bibitem[\protect\citeauthoryear{Sun, Feng, and Saenko}{2016}]{sun2016return}
Sun, B.; Feng, J.; and Saenko, K.
\newblock 2016.
\newblock Return of frustratingly easy domain adaptation.
\newblock In {\em AAAI}, volume~6, ~8.

\bibitem[\protect\citeauthoryear{Teerapittayanon, McDanel, and
  Kung}{2016}]{teerapittayanon2016branchynet}
Teerapittayanon, S.; McDanel, B.; and Kung, H.
\newblock 2016.
\newblock Branchynet: Fast inference via early exiting from deep neural
  networks.
\newblock In {\em ICPR},  2464--2469.
\newblock IEEE.

\bibitem[\protect\citeauthoryear{Tzeng \bgroup et al\mbox.\egroup
  }{2014}]{tzeng2014deep}
Tzeng, E.; Hoffman, J.; Zhang, N.; Saenko, K.; and Darrell, T.
\newblock 2014.
\newblock Deep domain confusion: Maximizing for domain invariance.
\newblock {\em arXiv preprint arXiv:1412.3474}.

\bibitem[\protect\citeauthoryear{Tzeng \bgroup et al\mbox.\egroup
  }{2017}]{tzeng2017adversarial}
Tzeng, E.; Hoffman, J.; Saenko, K.; and Darrell, T.
\newblock 2017.
\newblock Adversarial discriminative domain adaptation.
\newblock In {\em CVPR}, volume~1,  4--13.

\bibitem[\protect\citeauthoryear{Yuan, Yang, and Zhang}{2017}]{yuan2017hard}
Yuan, Y.; Yang, K.; and Zhang, C.
\newblock 2017.
\newblock Hard-aware deeply cascaded embedding.
\newblock In {\em ICCV},  814--823.
\newblock IEEE.

\end{thebibliography}
\bibliographystyle{aaai}
\end{document}